\documentclass[preprint,5p,times,twocolumn]{elsarticle}

\usepackage{amssymb}
\usepackage{fancyhdr}
\usepackage{hyperref}

\usepackage{color}

\newcommand{\newtext}[1]{\textcolor{black}{#1}}
\newcommand{\newtextrev}[1]{\textcolor{black}{#1}}

\journal{Computer Vision and Image Understanding}

\begin{document}

\begin{frontmatter}



\title{Recovering hard-to-find object instances\\ by sampling context-based object proposals}


\author{Jos{\'e} Oramas M.}
\author{Tinne Tuytelaars}



\address{KU Leuven, ESAT-PSI, iMinds\\
Kasteelpark Arenberg 10 - bus 2441\\
B-3001 Heverlee, Belgium}

\begin{abstract}
    In this paper we focus on improving object detection 
    performance in terms of recall. 
    We propose a post-detection stage during which
    we explore the image with the objective of recovering missed 
    detections. This exploration is performed by sampling object 
    proposals in the image.
    We analyze four different strategies to perform this 
    sampling, giving special attention to strategies
    that exploit spatial relations between objects.
    In addition, we propose a novel method to discover 
    higher-order relations between groups of objects.
    Experiments on the challenging KITTI dataset show that
    our proposed relations-based proposal generation strategies 
    can help improving recall at the cost 
    of a relatively low amount of object proposals.
\end{abstract}

\begin{keyword}
Object Detection \sep Object Proposal Generation \sep Context-based Reasoning \sep Relational Learning


\end{keyword}

\end{frontmatter}

\thispagestyle{fancy}

\section{Introduction}
\vspace{-2mm}
Object detection methods have become very effective at localizing 
object instances in images. Different methods have been proposed, 
ranging from methods that model the appearance of the object as it is 
projected on the 2D image space 
\cite{dalalTriggsCVPR05,felzenszwalbTPAMI2009,girshick14CVPR,HeSPPCNN} 
to methods that reason about physical properties of the objects in the 
3D scene \cite{Pepik3D2PM,ZiaCVPR14}. 
All these methods have one thing in common: they rely completely 
on appearance features, e.g. color, shape or texture, to describe 
the objects of interest. 
If the object is clearly visible in the image, appearance cues can be 
very strong.
Unfortunately, appearance-based approaches cannot cope well with 
more difficult cases, such as small object instances or highly 
occluded ones. In spite of some efforts in this direction 
(e.g. Frankenclassifier~\cite{Mathias2013Iccv}, Occlusion 
Patterns~\cite{PepikOcclusionPatterns}, Occlusion 
Boundaries~\cite{HoiemOcclusionBoundaries}), these mostly 
remain undetected, resulting in reduced recall.
In a real world setting, highly cluttered scenes and therefore 
small and occluded objects are actually quite common - probably 
more common than in typical benchmark datasets which are often 
object-focused (e.g. because they have been collected by 
searching images that have the object name mentioned in the tags).

\begin{figure}[h!]
\centering
\includegraphics[width=0.48\textwidth]{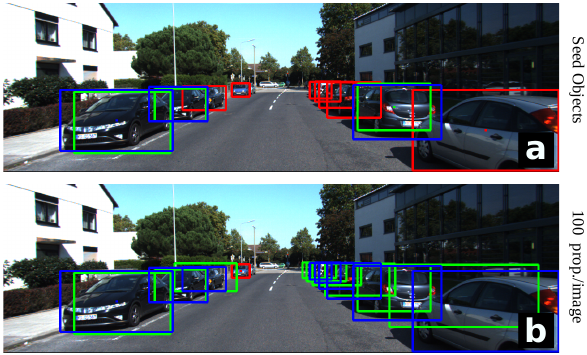}
\caption{ Object detections collected:
         a) after running a standard appearance-based detector, 
         b) after sampling 100 context-based object proposals post detection.
         Notice how we manage to recover many of the initially missed detections.
         Matched annotations are marked in blue, missed detections in red and 
         matching object proposals in green (Best viewed in color).
}
\label{fig:problemStatement}
\end{figure}

In recent years, several works 
\cite{choiLTW10,DesaiRF11_ijcv,HoiemPOPCVPR06,joramas:WACV14,perko2010a} 
have proposed the use of contextual information.  
These works typically follow a two-stage pipeline during testing. 
First, a set of detections is collected using an appearance-based 
detector. Then, using a pre-learned context model, out-of-context 
detections are filtered-out.
This strategy has been effective at improving object detection, 
specifically, in terms of precision. On the downside, objects 
missed by the object detector are not recovered, which leaves 
no room for improvement in terms of recall.
A possible explanation for this tendency, is that the high-precision 
low-recall area is often considered the more interesting part of 
the precision-recall curve \cite{AtanasoaeFPfaces,hongyuFPremoval}. 
Methods are optimized and typically perform well in this region. 
The high-recall low-precision area, on the other hand, receives 
little attention -- as if we all have come to accept there is some 
percentage of object instances that are just too hard to be found.
A very different view is common in the work on class-independent object proposals detection 
(e.g. objectness~\cite{AlexeObjectnessCVPR10,AlexeObjectnessPAMI12}, selective 
search~\cite{UijlingsIJCV2013}, edge boxes~\cite{ZitnickEdgeBoxes14}, 
deepProposals~\cite{Ghodrati_2015_ICCV} or deepBoxes~\cite{Kuo_2015_ICCV}).
When the object class is unknown, no-one expects a high precision, and it is only 
natural to focus on recall instead. A common evaluation protocol in this context is 
the obtained recall as a function of the number of window proposals per image. 
Here, we adopt the same evaluation scheme, but now for standard supervised object 
detection.

In a sense, this work is similar to \cite{Vedaldi09} which also focuses on 
recall instead of precision. The goal is to find as many 
object instances as possible, even if this comes at a cost, in the form of many 
false positives (low precision). Because of the lower precision, we refer to the 
detections as 'object proposals', as in the class-independent object detection work. 
This reflects the idea that further verification (e.g. using other modalities, other 
viewpoints or higher resolution imagery) may be required to separate the true 
positives from the many false positives  -- a process which may be application 
dependent and is out-of-scope of this work. 
We compare various strategies to generate object proposals: 
i) a sliding-window baseline, \
ii) two methods for class-independent object proposals (selective search 
\cite{UijlingsIJCV2013} and edge boxes \cite{ZitnickEdgeBoxes14}), and \
iii) two class-specific context-based schemes. 
Different from \cite{Vedaldi09}, which exploits intrinsic appearance features, 
we focus on context cues from other objects in the scene. 
Multiple objects in a scene often appear in particular spatial configurations. 
This means that detecting one object also provides information about possible locations of other 
objects. We start from a few high-confidence appearance-based detections and use these 
as seeds based on which other likely object locations  are identified. We explore 
one method that uses pairwise relations, and propose a new topic-based method that 
builds on higher-order spatial relations between groups of objects. 
We have found that, based on very simple features, relative location 
and pose, our method is able to discover arrangements 
between objects that resemble those found in the real world. 
Furthermore, it does not enforce restrictions on the number of objects 
participating in each of the higher-order relations.
For simplicity, we assume the ground plane to be known, 
both for the baselines as for the newly proposed context-based schemes.
However, note that if needed, these can be estimated in different ways 
(e.g. \cite{Bao2011,HoiemPOPCVPR06}).
We show that our method is able to bring significant improvement
to standard object detectors. For example, notice how in 
Fig.~\ref{fig:problemStatement}.b we manage to recover many of 
the initially missed detections (Fig.~\ref{fig:problemStatement}.a). 
This is achieved at a relatively low cost of just 100 additional 
object proposals.

The remainder of this paper is organized as follows: 
Sec.~\ref{sec:relatedWork} presents related work. 
In Sec.~\ref{sec:proposedMethod} we present the details of 
the analysis and of the methods for generating object proposals.
Experiments, results and discussions are presented in 
Sec.~\ref{sec:evaluation}. Then, Sec.~\ref{sec:limitations} 
addresses the current limitations of the proposed method.
Finally, Sec.~\ref{sec:conclusions} concludes this paper.

\section{Related Work}
\label{sec:relatedWork}

The analysis presented in this paper lies at the intersection of 
class-independent and context-based class-specific object detection. 
These two groups of work constitute the axes along which we position 
our work.

\subsection{Context-based class-specific object detection}

Contextual information, in the form of relations between
objects, has been successfully exploited to improve object 
detection performance in terms of precision 
\cite{choiLTW10,DesaiRF11_ijcv,felzenszwalbTPAMI2009,perko2010a}.  
However, objects missed by the object detector are not recovered. 
This, in consequence, leaves no room for improvement in terms of recall.
One work that tries to increase recall is the co-detection work of 
\cite{Bao_ECCV2012_codetection}. They exploit detections of the same object 
instances in multiple images to generate bounding boxes. 
Our work, on the other hand, operates on a single image.
Different from \cite{MottaghiCVPR14}, our method does not require an 
image segmentation step. Furthermore, our contextual models are defined 
in 3D space. This last aspect also separates our method from \cite{LiECCV14}. 
This makes our models easier to transfer to other datasets with other 
camera viewpoints and gives them some 
level of interpretability which can be exploited in other applications, 
e.g. autonomous driving or robotic manupulation. 
Different from \cite{YaoCVPR10}, 
which models human-object interactions, our context models are more flexible 
since the majority of the related objects may not occur in the scene, 
whereas in \cite{YaoCVPR10}, the objects and the parts of the body are always 
present.
Additionally, our work differs from 
\cite{choiLTW10,DesaiRF11_ijcv,felzenszwalbTPAMI2009,perko2010a} in
that we consider higher-order relations whereas most of the methods 
that exploit relations between objects focus on the pairwise case.
Recently, a small group of works \cite{CaoPureDependency14,joramas:WACV14,ZhangTrafficICCV13} 
that consider higher order relations has been proposed.
In \cite{CaoPureDependency14}, a Pure-Dependency \cite{pureDependency2011} 
framework is used to link groups of objects. In \cite{joramas:WACV14},
objects are grouped by clustering pairwise relations between them. The
work of \cite{ZhangTrafficICCV13} is able to reason about higher-order
semantics in the form of traffic patterns. Different from these works, 
our topic-based method to discover higher-order relations does
not require the number of participating objects to be predefined \cite{pureDependency2011}. 
Furthermore, objects do not need to be ``near'' in the space defined
by pairwise relations in order to be covered by the same higher-order
relation \cite{joramas:WACV14}. 
Finally, our method does not require scene-specific cues ( e.g. 
lane presence, lane width or intersection type), or
motion information \cite{ZhangTrafficICCV13}.

Another related work is \cite{KuttelCVPR10} where two methods 
are proposed to learn spatio-temporal rules of moving agents from video 
sequences. This is done with the goal of learning temporal dependencies 
between activities and allows interpretations on the observed scene.
Our method is similar to \cite{KuttelCVPR10} in that both 
methods perform spatial reasoning and both methods are evaluated 
in a street scene setting. 
Different from \cite{KuttelCVPR10} which aims at building scene-specific 
models, the models produced by our method are specific to the object 
classes of interest and not scene-dependent. Furthermore, while 
\cite{KuttelCVPR10} focuses more on motion (flow) cues, our method 
focuses on instance-based features (location \& pose).
Moreover, the method from \cite{KuttelCVPR10} requires video sequences 
and operates in the 2D image space while our method runs on still 
images and operates in the 3D space.

\subsection{Class-independent object detection} 

Another group of work operates under the assumption that there are regions 
in the image that are more likely to contain objects than others. Based 
on this assumption, the problem is then to design an algorithm to find 
these regions. Following this idea, \cite{AlexeObjectnessCVPR10} proposed 
a method where windows were randomly sampled over the image. Following
the sampling, a ``general'' classifier was applied to each of the windows.
This classifier relied on simple features such as appearance difference w.r.t.
the surrounding or having a closed contour and was used to measure the 
objectness of a window. In \cite{AlexeObjectnessCVPR10}, windows with 
high objectness are considered to be more likely to host objects.
Later, \cite{EndresProposalsPAMI14} proposed a similar method 
with the difference that their method generated object proposals from an 
initial segmentation step. This produced better aligned object proposals.
Similarly, \cite{UijlingsIJCV2013} proposed a selective search method 
which exploits the image structure, in terms of segments, to guide the 
sampling process. In addition, their method imposes diversity 
by considering segment grouping criteria and color spaces with
complementary properties. Recently, \cite{ZitnickEdgeBoxes14}
proposed a novel objectness measure, where the likelihood of a 
window to contain an object is proportional to the number of contours
fully enclosed by it.
A common feature of this group of work is that their precision is less 
critical. The number of generated proposals is anyway only a small percentage 
of the windows considered by traditional sliding window approaches.
On the contrary, these methods focus on improving detection recall by
guiding the order in which windows are evaluated by later class-specific
processes.
In \cite{AleHeeTeh2012a}, these ideas were integrated in a context-based 
detection setting where new proposals are generated sequentially based 
on previously observed proposals following a class-specific context model.
Inspired by these methods we propose to complement a traditional object 
detector with an object proposal generation step. The objective of this 
additional step is to improve detection recall even at the cost of more 
false positives. Different from \cite{AleHeeTeh2012a}, which just 
returns a single window per image (thus detecting a single object instance), 
we generate several windows per image with the objective of recovering 
as many object instances as possible. Moreover, our context information 
is object-centered.
 
\newtext{ 
Recently, \cite{long2013active} proposed ``location relaxation'', 
a two-stage detection strategy where candidate regions of the 
image are identified using coarse object proposals generated 
from bottom-up segmentations. Then, based on these proposals, 
a top-down supervised search is performed to precisely localize 
object instances.
Similar to \cite{long2013active} we propose a two-stage strategy 
to improve object detection. However, instead of focusing on 
refining object localization our focus is on maximizing the number 
of detected instances. In this aspect, the proposed method and 
the work from \cite{long2013active} complement each other since 
the proposed method can be used to coarsely localize object instances 
while the strategy from \cite{long2013active} can be used to 
improve localization accuracy. Another difference, is that while 
the method from \cite{long2013active} uses “local” class-specific 
models to improve the localization of a specific object instance, 
our method uses context models to explore candidate locations 
of other instances.
}

\section{Proposed Method}
\label{sec:proposedMethod}
The proposed method can be summarized in 2 steps: In a first stage, 
we run a traditional object detector which produces a set of object detections.
Then, in a second stage, we sample a set of object proposals aiming to recover
object instances possibly missed during the first stage.

\subsection{Class-specific object detection}
\label{sec:detection}

The main goal of this work is to recover missed object instances 
after the initial detection stage has taken place. Given this focus on
the post-detection stage, for the object detection stage we start 
from an off-the-shelf detector.
In practice, given a viewpoint-aware object detector, i.e. a detector 
that predicts the bounding box and viewpoint of object instances, we collect 
a set of 2D object detections $o=\{o_1,o_2,...,o_n\}$ where each
object detection $o_i=(b_i,\alpha_i,s_i)$ is defined by its detection 
score $s_i$, its predicted viewpoint $\alpha_i$ and its 2D bounding 
box coordinates $b_i=(x_{1i},y_{1i},x_{2i},y_{2i})$.

\subsection{Object Proposal Generation Methods}
\label{sec:objectProposalGeneration}

Traditional appearance-based object detectors have proven to be effective
to detect objects $o$ with high confidence for the cases when objects of 
interest are clearly visible. At the same time, for small or highly-occluded 
object instances its predictions are less reliable resulting in a significant
number of object instances being missed. To overcome this weakness we propose, as
a post-detection step, to sample (class-specific) object proposals $o'$ with 
the goal of recovering missed detections. 
We analyze four strategies to generate these proposals, as discussed in
the next four sections.

\subsubsection*{Relaxed Score Detector}
A first, rather straightforward method to recover missed detections
consists of further reducing the threshold $\tau$ used as cutoff in 
the object detector. This is a widely used strategy, even though it 
usually does not increase recall that much. 
\newtext{
We refer to this strategy as {\em Relaxed Score Detector}. This 
strategy consists of the original object detector with non-maximum 
suppression (NMS) performed with default settings while reducing 
drastically the threshold $\tau$ for the detection score.
}

\subsubsection*{Relaxed NMS Detector}
\newtext{
In addition, we define an alternative strategy to relax the object 
detector. Instead of lowering the threshold $\tau$, we remove the 
non-maximum suppression step present in most object detectors (including the one used in our experiments). 
For a given threshold value, this results in many more object proposals 
being generated. This allows to detect objects highly-occluded by 
other objects of the same class.
We refer to this strategy as {\em Relaxed NMS Detector}.
}

\subsubsection*{3D Sliding Window}
This is a {\footnotesize 3D} counterpart of the {\footnotesize 2D} sliding window approach used
by traditional detectors (e.g. \cite{felzenszwalbTPAMI2009}). This approach is 
inspired by the work of Hoiem et al.~\cite{HoiemPOPCVPR06}. We
assume the existence of a ground plane that supports the objects of interest. 
Given the ground plane, we densely generate a set of {\footnotesize 3D} object proposals 
{\footnotesize $O'=\{O'_1,...,O'_m\}$} resting on it for each of the discrete
orientations {\footnotesize $\theta_k=\{\theta_1,...,\theta_K\}$}. 
Each {\footnotesize 3D} object proposal, {\footnotesize $O'=(X,Y,Z,L,W,H,\theta)$}, is defined by its {\small 3D} location 
{\footnotesize $(X,Y,Z)$}, its physical length, width and height {\footnotesize $(L,W,H)$} and its orientation 
$\theta$ in the scene. We define the length, width and height {\footnotesize $(L,W,H)$} of the 
proposed {\footnotesize 3D} object proposals {\footnotesize $O'$} as the mean values of annotated {\footnotesize 3D} objects in 
the training set. We drop the {\footnotesize 3D} location coordinate {\footnotesize $Y$} since all the {\footnotesize 3D} object proposals are 
assumed to be supported by the ground plane, hence {\footnotesize $Y=0$} for all the proposals.
Then, once we have generated all the {\footnotesize 3D} objects that can physically be in the 
scene, using the camera parameters we project each of the {\footnotesize 3D} object proposals
{\footnotesize $O'$} to the image space, assuming a perspective camera model, producing a set of
{\footnotesize 2D} object proposals $o'$. Specifically, each {\footnotesize 2D} proposal $o'$ is obtained by 
projecting each of the corners of the {\footnotesize 3D} proposal {\footnotesize $O'$}, and selecting the 
{\footnotesize 2D} points that enclose the rest.
Note that due to the box representation, objects with opposite orientations 
(orientation difference = {\footnotesize 180}$^{\circ}$) will project onto the same {\footnotesize 2D} bounding boxes.
For this reason we only generate proposals for a smaller set of {\footnotesize $K/2$} discrete 
viewpoints.

\subsubsection*{Class Independent Object Proposals}
 Here we follow the strategy of generic, class-independent, object proposal
 generators. A crucial part of this strategy is to define a proper objectness
 measure to be able to estimate how likely it is for a window defined over an image to
 contain an object of any class. In this analysis we evaluate the effectiveness
 of this strategy to recover missed detections. Particularly, 
 we use the Selective Search \cite{UijlingsIJCV2013} 
 and Edge Boxes \cite{ZitnickEdgeBoxes14} methods. 
 See \cite{Hosang2014Bmvc} for a benchmark of methods for detecting 
 class independent object proposals.

\begin{figure*}[ht!]
\centering
\includegraphics[width=0.66\textwidth]{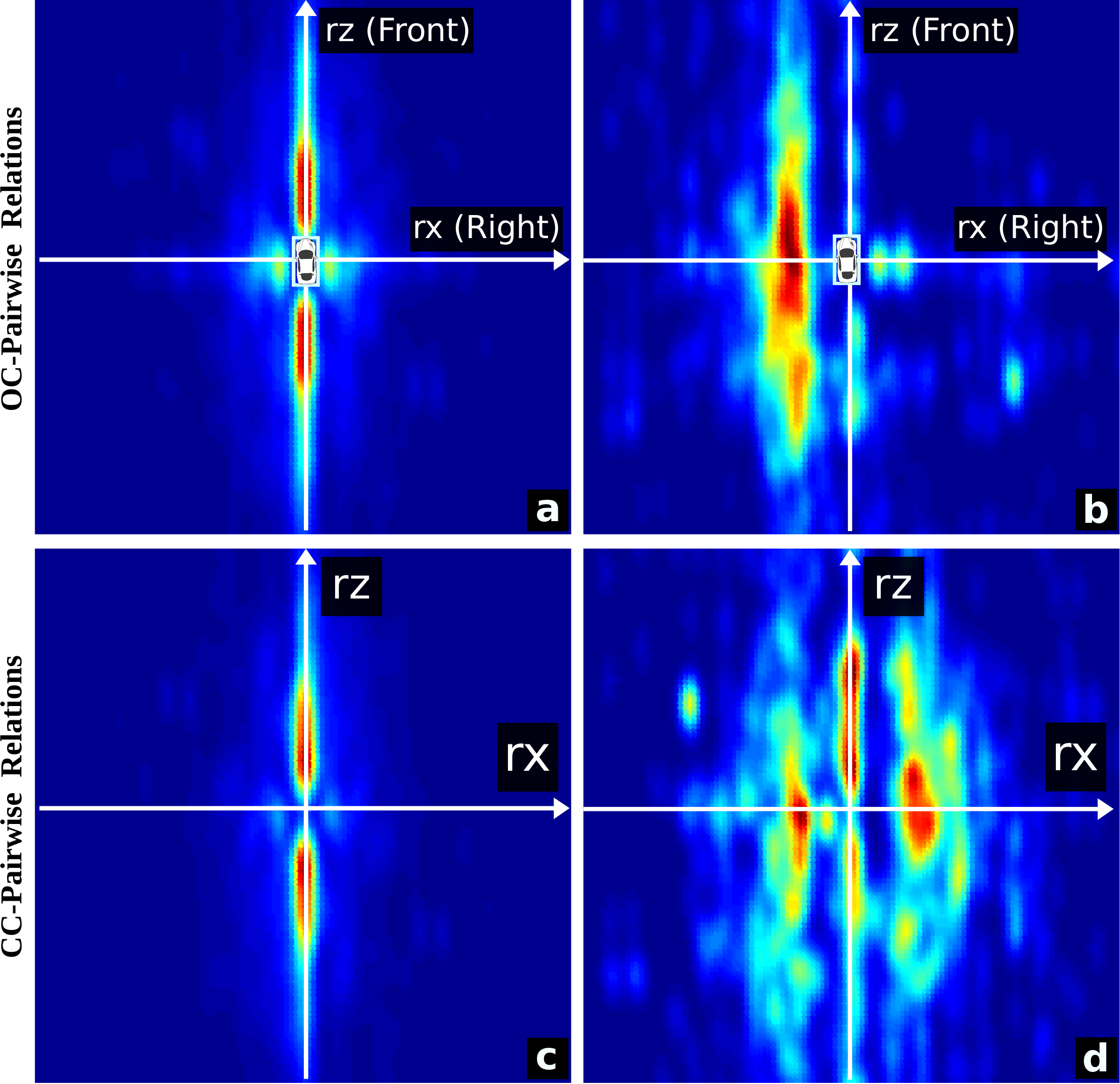}
\caption{ Distribution of pairwise relations for cars with the same pose (a,c) and opposite pose (b,d) respectively. 
	  Top row corresponds to object-centered (OC) relations while the bottom row corresponds to camera-centered 
	  (CC) relations.}
\label{fig:pairwiseRelations}
\end{figure*}

\subsection{Class-specific Context-based Object Proposals}
\label{sec:contextBasedProposals}
In this strategy we generate a set of object proposals $o'$ as a 
function {\small $o'=f^\eta(o)$} of the object detections $o$ predicted by the 
appearance-based detector. The function $f^\eta$ enforces contextual 
information in the form of relations between object instances. This way, 
all the proposals $o'$ sampled from $f^\eta$ follow a distribution of 
relations previously seen in the training data where $\eta$ is the number 
of object instances participating in the relation. This produces a 
relation-driven search where given a seed object $o_i$ object proposals 
$o'$ are sampled at locations and with poses that satisfy these relations. 
In this paper we propose two relation-driven functions: $f^2$ for 
the case of objects being associated by pairwise relations, and 
$f^+$ for the case when objects are associated by higher-order relations.

\paragraph{From {\footnotesize 2D} object detections to {\footnotesize 3D} objects in the scene}
In this work, reasoning about relations between objects is performed in the 3D scene. 
For this reason, we first need to project the object detections used 
as seeds on the {\footnotesize 3D} scene using the groundplane. We define the objects 
{\footnotesize $O=\{O_1,O_2,...,O_n\}$} as 
3D volumes that lie within this {\footnotesize 3D} space. 
Each object {\footnotesize $O_i=(X_i,Y_i,Z_i,L_i,W_i,H_i,\theta_i,s_i)$},
is defined by its {\footnotesize 3D} location coordinates {\footnotesize $(X_i,Y_i,Z_i)$}, 
its size {\footnotesize $(L_i,W_i,H_i)$}, its pose $\theta_i$ in the {\footnotesize 3D} 
scene and its confidence score $s_i$. 
We assume that all the objects rest on a common ground plane, 
so {\footnotesize $Y_i=0$} for all the objects. For brevity, we drop 
the {\footnotesize $Y$} term, then each object is defined as 
{\footnotesize $O_i=(X_i,Z_i,L_i,W_i,H_i,\theta_i,s_i)$}.
In order to define the set of {\footnotesize 3D} objects {\footnotesize $O$} from the set of {\footnotesize 2D} 
objects $o$, we execute the following procedure: first, given a set of annotated
3D objects, we obtain the mean size (length,width and height) of the
objects in the dataset. Second, assuming a calibrated camera, we densely
generate a set of 3D object proposals {\footnotesize $O'$} over the ground plane, very similar the {\footnotesize 3D} Sliding Windows method from Sec.~\ref{sec:objectProposalGeneration} .
Third, each of the {\footnotesize 3D} object proposals from {\footnotesize $O'$} is projected in the 
image plane producing a set of {\footnotesize 2D} proposals $o'$. Then, for each 
object detection $o_i$ we find its corresponding proposal $o_i'$
by taking the proposal with highest intersection over union score, as proposed in Pascal VOC Challenge~\cite{PascalVOC2012}.
Finally, we use the {\footnotesize 3D} location {\footnotesize $(X_i',Z_i')$} from the {\footnotesize 3D} proposal {\footnotesize $O_i'$} from 
which $o_i'$ was derived and the viewpoint angle $\alpha_i$, predicted
by the detector, to estimate the pose angle $\theta_i$ of the object {\footnotesize $O_i$}
in the scene. As a result, we obtain a set of {\footnotesize 3D} objects defined as  
{\footnotesize $O_i=(X_i,Z_i,L_i,W_i,H_i,\theta_i,s_i)$}.

\begin{figure*}[t!]
\centering
\includegraphics[width=1\linewidth]{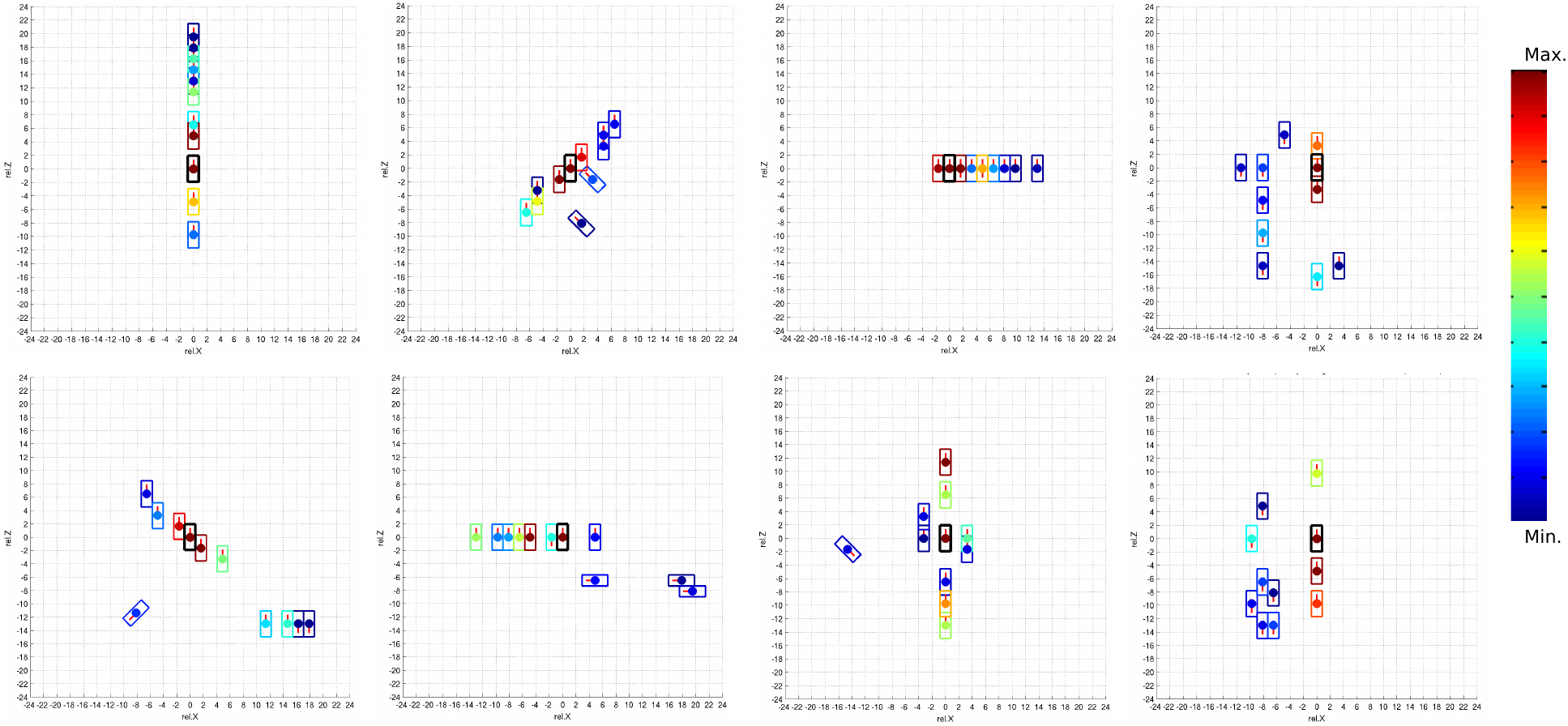}
\caption{Some of the discovered relational topics from an object-centered perspective. 
         For each topic, the reference object is in the center and colored in black. 
         The related objects are presented with their occurrence likelihood color-coded in 
         jet scale. Notice how the discovered topics resemble traffic scenarios from urban
         scenes. 
         For visualization purposes, each object is being plotted
         with average size of the annotations in the training set of images. 
         We only show the top 10 most likely words per topic.
         (Best viewed in color).}
\label{fig:discoveredTopics}
\end{figure*}

\paragraph{Pairwise Relations~($f^2$)}
Pairwise relations between 3D objects are computed as proposed in \cite{joramas:ICCV13}. Following the procedure from \cite{joramas:ICCV13}  we define \textit{camera-centered~(CC)} pairwise relations 
by centering the frame of reference in the camera.
Then, from this frame of reference we measure relative location 
and orientation values between object instances. 
Alternatively, we define \textit{object-centered~(OC)}
relations in which, first, we center the frame of reference on 
each of the object instances and then measure the relative 
values between them. As a result, we obtain a set of relations 
{\footnotesize $R$} for each image. Each pairwise relation $r_{ij}$ is defined 
as {\footnotesize $r_{ij}=(r_X,r_Z,r_\theta)$}, where {\footnotesize $(r_X,r_Z)$} 
represent the relative location of the object and $r_\theta$ represents
the relative pose between the object instances. 
We compute pairwise relations between each pair of objects 
within each image of the training set. Then, using kernel 
density estimation (KDE) we model the distribution {\footnotesize $p(r_{ij})$}.
This is a simple method that manages to find some common arrangements
in which pairs of objects co-occur. See Fig.~\ref{fig:pairwiseRelations} 
for some examples.
During the proposal generation stage, we sample a set of relations 
$r'$ from this distribution. Then, for each seed object {\footnotesize $O$} we 
generate object proposals {\footnotesize $O'$} following the sampled relations $r'$.
Finally, the {\footnotesize 3D} object proposals {\footnotesize $O'$} are projected into the image 
plane producing the {\footnotesize 2D} object proposals $o'$.

\paragraph{Higher-order Relations Discovery~($f^+$)}
Given a set of training images containing objects occurring in a scene,
our goal is to discover the underlying higher-order relations that 
influence the location and orientation in which each object 
instance occurs w.r.t. each other.
A similar problem, of discovering abstract topics $t=\{t_1,t_2,...,t_T\}$
that influence the occurrence of words $w$ within a document $d$,
is addressed by Topic Models \cite{BleiLDA2003} \cite{griffiths:TopicModels}. 
Motivated by this similarity we formulate our higher-order relation 
discovery problem as a topic discovery problem.
According to the topic model formulation, a document $d_i$ can
cover multiple topics $t_k$ and the words $w$ that appear
in the document reflect the set of topics $t_k$ that it covers.
From the perspective of statistical natural language processing,
a topic $t_k$ can be viewed as a distribution over words $w$;
likewise, a document $d$ can be considered as a probabilistic
mixture over the topics $t$. 

In order to meet this formulation in our particular setting,
given a set of training images, we first compute pairwise relations $r_{ij}$ 
between all the objects {\footnotesize $O_i$} within each image as before. 
Then, for each object {\footnotesize $O_i$} we define a document $d_i$ where 
the words $w$ are defined by the pairwise relations $r_{ij}$ 
that have the object {\footnotesize $O_i$} as the source object. 
Additionally, we experiment with an alternative way to compute 
the pairwise relations between objects. Specifically, we run 
tests with a variant of the relative pose attribute of the relation 
where instead of considering the pose of the target object we 
consider the orientation of its elongation only (similar to 
\cite{joramas:BMVC14} ). This orientation is less affected by 
errors during prediction, since traditional pose estimators 
tend to make mistakes by confusing opposite orientations, e.g.  
front-back, left-right, etc.

In order to make the set of extracted pairwise relations {\footnotesize $R$}
applicable within the topic model formulation we quantize them into 
words (although word-free topic models have been proposed as well 
\cite{RematasNIPSWS12}). To this end, we discretize the space defined 
by the relations {\footnotesize $R$} by {\footnotesize $(W/2,W/2,\theta_d)$}
where {\footnotesize $W$} is the average width of the annotated 3D objects 
in the training set, and $\theta_d$ is a predefined number
of discrete poses of the object, 8 in our experiments. 
At this point, we are ready to perform topic modelling in our data.
Here we use Latent Dirichlet Allocation \cite{BleiLDA2003} for 
topic modelling. For inference, we follow a Gibbs sampling method 
as in \cite{griffiths:TopicModels}.
%
Our main goal is to identify the set of topics $t$ that define
higher-order arrangements between objects {\footnotesize $O$} in the scene.
In our experiments we extract 16 topics from our documents $d$.
Fig~\ref{fig:discoveredTopics} shows a top view of a subset of 
the discovered topics when considering object-centered pairwise 
relations as words. Notice how some of the topics resemble common 
traffic patterns of cars in urban scenes. These topics represent
the underlying higher-order relations that we claim influence the
way in which objects tend to co-occur. 

During the object proposal generation stage, we assume that each
3D object {\footnotesize $O_i$}, estimated from the seed object detection $o_i$,  
is related with the object proposals {\footnotesize $O'$} under higher-order 
relations. For simplicity, we assume that all the higher-order 
relations (topics) are equally likely to occur. Object proposals {\footnotesize $O'$} are then 
generated by sampling the word distributions $p(w|t)$ given each 
of the topics $t$. Finally the sampled 3D object proposals {\footnotesize $O'$}
are projected to the image plane, yielding $o'$.
The assumptions made at this stage have three desirable effects. 
First, object proposals are sampled in such a way that they follow
the higher-order relations between objects. Second, the exploration
process gives priority to the most likely proposals from each of the 
discovered higher-order relations, see Fig.~\ref{fig:discoveredTopics}.
Third, we are able to reason about higher-order relations 
even for the scenario when just one object detection $o_i$ 
was collected by the detector.


\begin{figure*}[t!]
\centering
\includegraphics[width=0.86\linewidth]{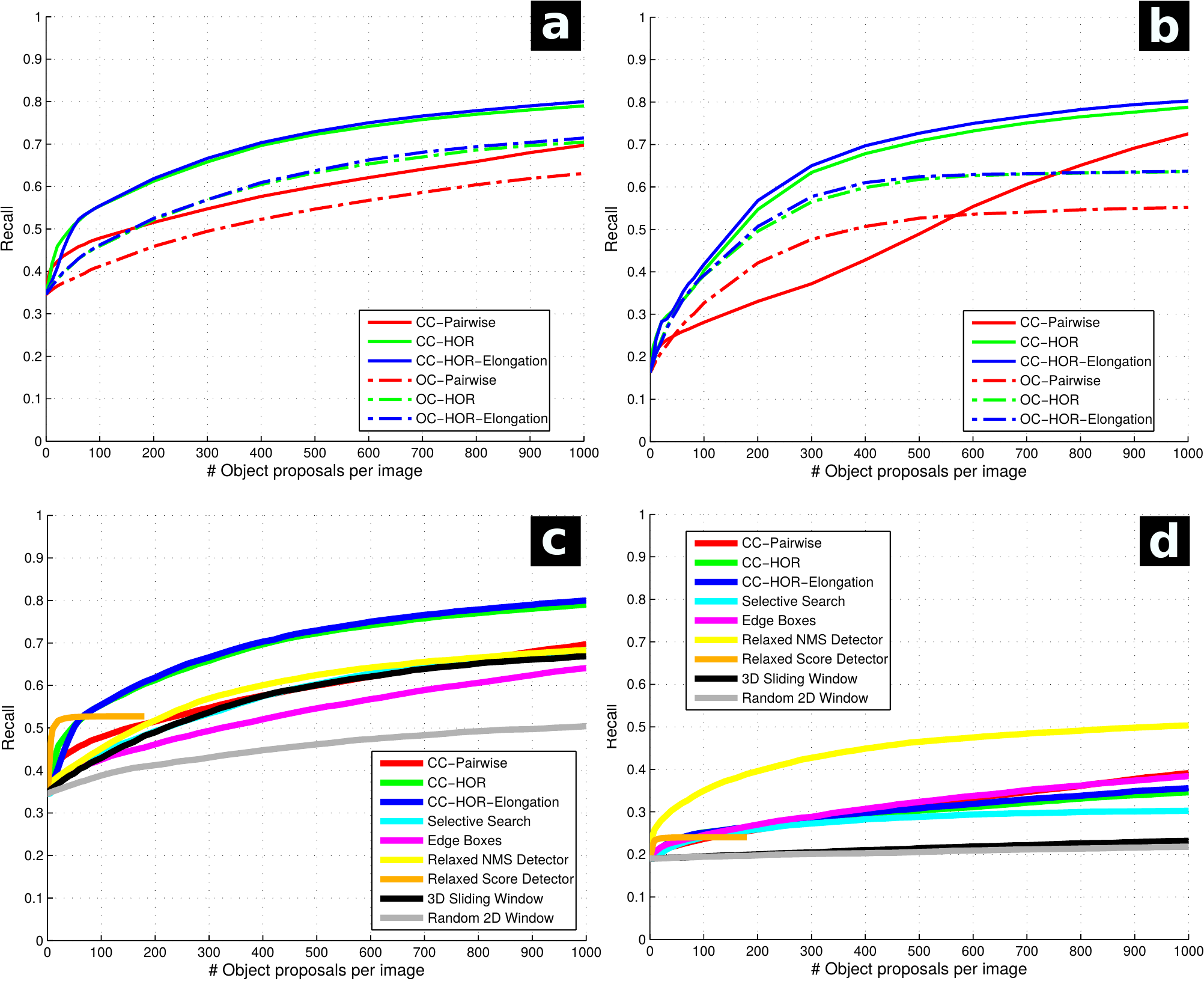}
\caption{Recall vs. number of generated proposals for our Relations-based methods 
a) when all the object detections reported by the detector are used as seed objects,  
and b) when only the top scoring object detection reported by the detector is used as seed 
object. Comparison with non-contextual strategies using: c) a traditional matching 
criterion (IoU$>$0.5), and d) using a stricter matching criterion (IoU$>$0.75).}
\vspace{-3mm}
\label{fig:resultsPlots}
\end{figure*}

\section{Evaluation}
\label{sec:evaluation}
\noindent\textbf{Experiment details:}
We perform experiments on the {\footnotesize KITTI} object detection benchmark~\cite{Geiger12KITTI}.
This dataset constitutes a perfect testbed for our analysis since it covers a wide
variety of difficult scenarios ranging from object instances with high occlusions 
to object instances with very small size. Furthermore, it provides precise annotations
for objects in the 2D image and in the 3D space, including their respective viewpoints
and poses. 
\newtext{In our experiments, and in contrast to standard procedure on the KITT object 
detection benchmark~\cite{Geiger12KITTI}, we consider all the object instances occurring 
on the images independently of their size, level of occlusion and/or truncation.}
Since this is a benchmark dataset, annotations are not available for the test set.
For this reason, we focus our experiments on the training set. Using the time stamps
of the dataset, we split the data into two non-overlapping subsets of equal size. 
The first subset is used for training, the second subset is used to evaluate the performance 
of our method. We focus on cars as the class of interest given its high occurrence within 
this dataset which makes it appropriate for reasoning about relations between objects. 
We focus our evaluation on images with two or more objects, where it is possible to
define such relations. This leaves us with two subsets consisting of 2633 images each 
that are used for training and testing, respectively.
Matching between annotated objects and object proposals is evaluated based on the 
intersection over union (IoU) criterion from Pascal {\footnotesize VOC} \cite{PascalVOC2012}.
We report as evaluation metric the recall as a function of the number of object proposals 
generated per image, as is often used for evaluating object proposal methods.
In this analysis we use mainly the {\footnotesize LSVM-MDPM-sv} detector from \cite{GeigerNIPS11} to collect
the initial set of object detections. {\footnotesize LSVM-MDPM-sv} is an extension of the 
Deformable Parts-based model ({\footnotesize DPM}) detector~\cite{felzenszwalbTPAMI2009} where 
a component is trained for each of the discrete object viewpoints to be predicted. 
In this case, {\footnotesize LSVM-MDPM-sv} is trained to predict eight viewpoints.

As baselines we use the \textit{Relaxed Score Detector}, the \textit{Relaxed NMS Detector}, 
the \textit{3D Sliding Window} proposals, the proposals generated by \textit{Selective Search} 
\cite{UijlingsIJCV2013} and \textit{Edge Boxes} \cite{HeSPPCNN}.  
For the case of class-specific context-based proposals,  
we evaluate one method based on pairwise relations, \textit{Pairwise},
and two methods based on higher-order relations, \textit{HOR} and \textit{HOR-Elongation},
where the latter is the variant based on object elongation orientation instead
of object pose.
For all the context-based strategies, for the special case when no seed objects are available, 
i.e. images where the object detector was unable to find detections above 
the threshold (12$\%$ of the images) , we fallback to the \textit{3D Sliding Window} strategy 
and consider the proposals proposed by this strategy for that image.
We evaluate the changes
in performance when considering camera-centered (\textit{CC}) relations vs. 
object-centered (\textit{OC}) relations.


\subsection*{Exp.1: Relations-based Object Proposals}
In this first experiment we focus on evaluating the strategies based
on relations between objects. 
We consider as seed objects for our strategies the object detections collected 
with the detector \cite{GeigerNIPS11}.
Fig.~\ref{fig:resultsPlots}.a presents performance on 
the range of {\small [0,1000]} generated object proposals.

\begin{figure*}
\centering
\includegraphics[width=1\linewidth]{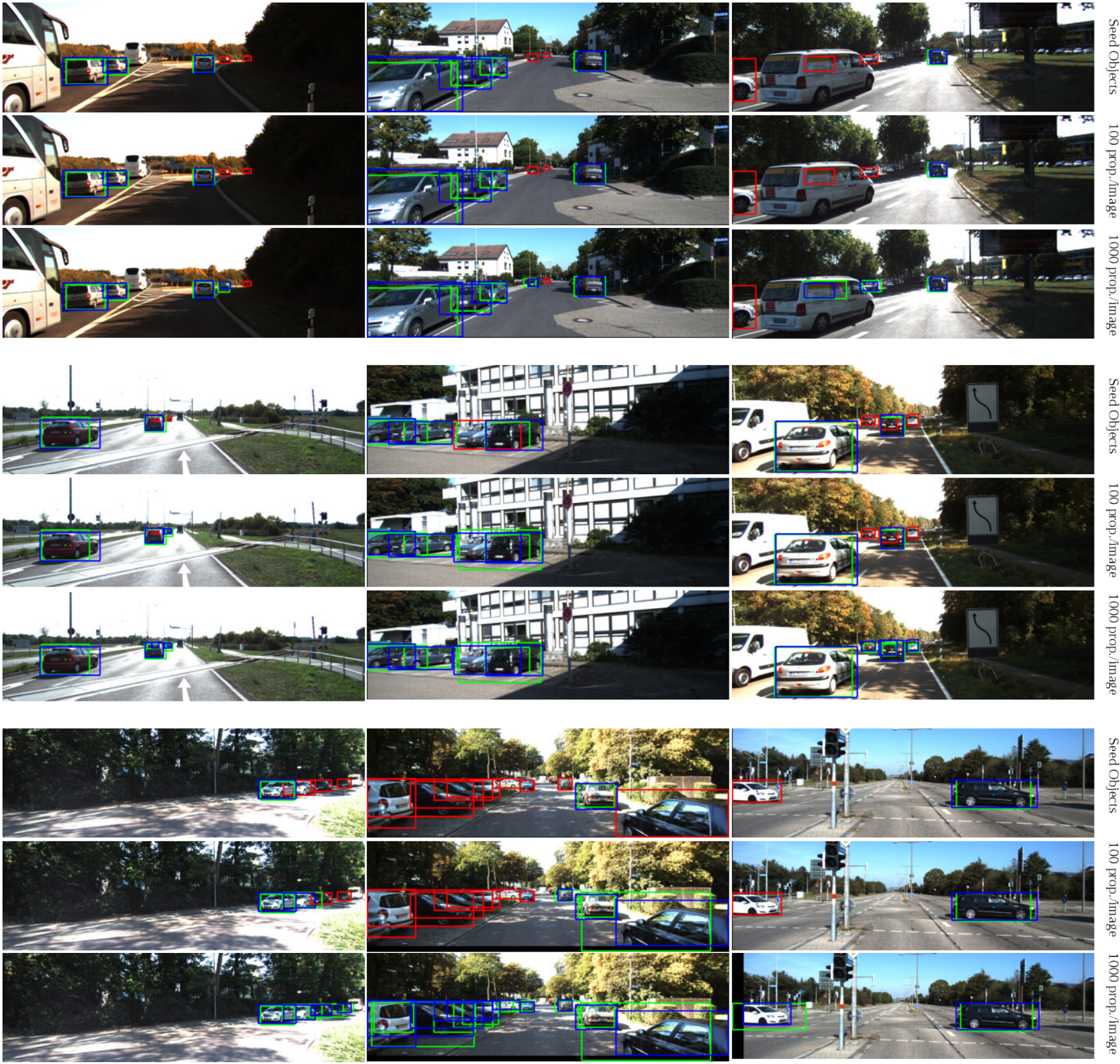}
\caption{Object proposals generated in chronological order using the context-based
          strategy based on camera-centered higher-order relations.
	  Matched object annotations are marked in blue, missed detections are marked
	  in red and matching object proposals are color-coded in green.
          First row, seed objects collected with the object detector \cite{GeigerNIPS11};  
          second row, results after sampling 100 object proposals; and  
          third row, results after sampling 1000 object proposals
         (Best viewed in color).}
\label{fig:qualitativeExamples}
\end{figure*}

\textit{Discussion:} 
Strategies based on \textit{CC}
higher-order relations seem to dominate the results.
They achieve around 10\% higher recall than all other methods
over a wide range of the curve.
This can be attributed to the fact that higher-order relations 
consider object arrangements with more than two participating
objects. This allows them to spot a larger number of areas
that are likely to contain objects. In addition, higher-order
relations cover a wider neighborhood, whereas the pairwise relations 
have a more ``local'' coverage (i.e. they explore mostly 
a small neighborhood around the seed detections). As a result, 
strategies based on higher-order relations are able to explore 
a large part of the image.
This is more visible in the range {\small [0,500]} of the sampled proposals,
where recall from methods based on higher-order relations increases faster
than for pairwise relations. 
This can be further verified in Fig.~\ref{fig:qualitativeExamples}.
A deeper inspection of the qualitative results (Fig.~\ref{fig:qualitativeExamples})
produced by our methods reveals a particular trend on how it addresses 
object instances of different sizes. Our method first focuses on objects in the 
3D vicinity of the seed objects, i.e. with similar projected 2D size. Eventually, objects with 
different 2D sizes to the seeds are explored.

Further we note  that strategies based on \textit{CC} relations 
have superior performance compared to their \textit{OC} counterparts.
This can be partly attributed to the fact that proposals sampled 
following \textit{OC} relations are affected by errors during the 
prediction of the pose of the seed object.
Moreover, the camera setup in the {\small KITTI} dataset is fixed, introducing 
low variability in the \textit{CC} relations.
In a scenario with higher variability on camera viewpoints we expect 
\textit{OC} relations to have superior performance over 
\textit{CC} relations.
In addition, for the case of \textit{CC} relations,
the higher-order relations where the elongation orientation 
is considered are slightly better, albeit only marginally so.
This can be attributed to the fact that the orientation of the elongation
of an object is less affected by errors in the pose estimation. 
Moreover, by defining \textit{CC} relations we also avoid the 
noise introduced in the pose of the seed objects.

Despite the difference in performance between the proposed strategies,
it is remarkable that we are able, on average, to double the initial 
recall obtained by the object detector by following relatively simple
strategies. This suggests that object proposal generation should not be 
employed solely as a pre-detection step as it is commonly found in the 
literature \cite{AlexeObjectnessCVPR10,EndresProposalsPAMI14,UijlingsIJCV2013,ZitnickEdgeBoxes14}. 
Furthermore, this suggests that there is some level of interoperability
between object detection and object proposal generation methods.


\subsection*{Exp.2: Starting from a Single Object Seed}
This experiment is similar to the previous experiment with the difference 
that for each image we only consider the top scoring object detection as
seed object.
As stated earlier, appearance-based detectors can be reliable at levels of 
high precision and low recall. The objective of this experiment is to measure
what performance can be achieved if we start from the most reliable seed object only.
Similar to the previous experiment, Fig.~\ref{fig:resultsPlots}.b
shows performance on the range of {\small [0,1000]} generated object proposals.

\textit{Discussion:} 
A quick inspection of Fig.~\ref{fig:resultsPlots}.b 
shows similar trends as the ones observed in the previous experiment. 
However, different
from the previous experiment, recall is relatively lower in the range
of {\small [0,100]} proposals. This is to be expected since we start from a 
smaller pool of seed objects. However, it is surprising to see how we
can achieve nearly similar 
performance from 400 proposals upwards by just starting from a 
single seed object. This further supports the idea of interoperability 
between object detectors and object proposal generators.

\subsection*{Exp.3: Comparison with non-contextual strategies}
Next, we compare the performance of the 
relations-based strategies w.r.t. the non-contextual methods of 
Section~\ref{sec:objectProposalGeneration}.
We consider as relation-based strategies the \textit{CC} variants 
only since, in the previous experiments, they achieved higher performance 
than their \textit{OC} counterparts.
As non contextual strategies we consider the \textit{Relaxed Score Detector}, 
the \textit{Relaxed NMS Detector}, the \textit{3D Sliding Window}, 
\textit{Selective Search} \cite{UijlingsIJCV2013}, 
and \textit{Edge Boxes}~\cite{ZitnickEdgeBoxes14}.
We report results considering all the detections as seed 
objects in Fig.~\ref{fig:resultsPlots}.c.

\textit{Discussion:}
We notice that the contextual strategies based on higher-order relations
have a superior performance than all the other strategies.
Interestingly, a clear difference can be noted between the performance of
contextual and non-contextual strategies. Except for the 
\textit{Relaxed Score Detector}, in the range of {\small [0,200]}, all 
the contextual strategies achieve superior performance than the non-contextual
counterparts. This suggests that indeed contextual information is useful 
for an early exploration of regions of the image that are likely to host 
instances of the objects of interest.
\newtext{
By  observing the performance of the 'Relaxed' versions of our local detector, 
we can verify the effect that the score threshold and NMS steps have on the 
obtained recall. As can be noted in Fig.~\ref{fig:resultsPlots}.c, when 
reducing the detection score threshold (\textit{Relaxed Score Detector}), the 
set of hypotheses predicted for each image is much lower ($<$200 per image) 
than when the NMS step is reduced (\textit{Relaxed NMS Detector}). 
Due to its stricter NMS step, the \textit{Relaxed Score Detector} produces less 
overlapping hypotheses, hence performing a faster exploration of the image space. 
This is evident since it reaches relatively high recall ($\sim$0.5) at the cost 
of less than 50 proposals per image. On the downside, due to its limited number 
of predicted hypotheses, this recall is not able to increase significantly. On 
the contrary, when the NMS step is removed, the \textit{Relaxed NMS Detector} reaches 
the 0.5 recall of the \textit{Relaxed Score Detector} at $\sim$200 proposals per 
image and is able to reach up to a recall of 0.7 later on the curve. 
}
%

\subsection*{Exp.4: Proposal Localization/Fitting Quality}
In this experiment we measure the quality of the object 
proposal to localize and fit the region of the recovered object instance. 
For this purpose, in this experiment we employ a stricter matching 
criterion \cite{PascalVOC2012} of at least 0.75 IoU 
between the bounding boxes of the object annotations and the object proposals, 
respectively.
This is inspired by \cite{Hosang2014Bmvc}, where it is claimed that 
a 0.5 IoU is insufficient for evaluating object proposals.
We evaluate the performance of the same, contextual and non-contextual, 
strategies from Exp.3.
In Fig.~\ref{fig:resultsPlots}.d we report results 
considering all the detections as seed objects with stricter IoU measure.

\textit{Discussion:}
Recall values obtained in this experiment are significantly reduced 
now that matching an object is a more complicated task.
The performance of the \textit{Relaxed NMS Detector} is surprisingly high.
This can be attributed to the fact that with the non-maximum suppression
step removed the \textit{Relaxed NMS Detector} is able
to exhaustively explore the areas where appearance has triggered 
a detection.
\newtext{This is further confirmed when comparing its performance 
with the one of the \textit{Relaxed Score Detector}. Due to its stricter
NMS step, the \textit{Relaxed Score Detector} is not able to explore 
possible bounding box variants occurring on a candidate region, thus 
resulting in poorer hypothesis bounding box matching.
}
In addition, we notice that pairwise relations are now outperforming
the higher-order alternatives in the range of {\footnotesize [500,1000]} 
proposals/image. This may be caused by the discrete nature of the words in the 
topic models which are used to discover higher-order relations.
As a result, the proposals generated from higher-order relations are
spatially sparser than the ones produced by pairwise relations.
The strategies based on pairwise relations tend to first concentrate 
on regions of high density before exploring other areas. This is 
why we notice improvements in the range {\footnotesize [500,1000]} and not earlier.
These observations hint at a possible weakness of our relation-based
strategies to generate object proposals. On one hand, relation-based
proposals have some level of sparsity embedded, in our case, either
by vector quatization of the relational space or by assuming mean 
physical sizes for the objects in the scene, when reasoning in {\footnotesize 3D}.
This can be a weakness compared to the exhaustive \textit{Relaxed NMS Detector}
strategy, when the objective is to have fine localization.
On the other hand, relation-based strategies seem to be better suited
for ``spotting'' the regions where the objects of interest might be. 
This is supported by their superior recall in Exp.1.
This further motivates our idea of a joint work of object 
detectors and object proposal generators.

\subsection*{Exp.5: Measuring Detection Performance}
\begin{figure}[ht!]
\centering
\includegraphics[width=0.98\linewidth]{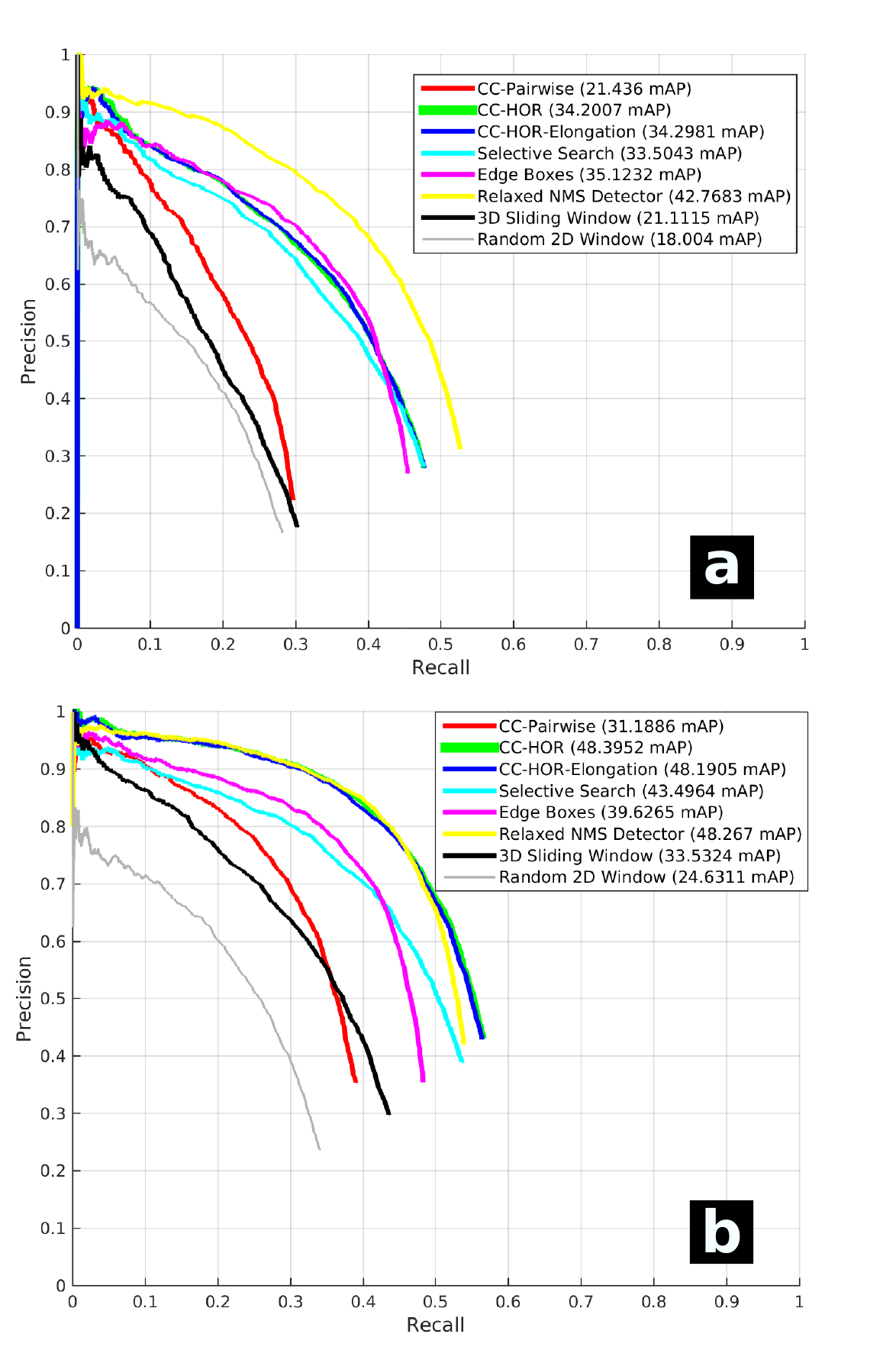}
\caption{Object detection Mean Average Precision (mAP) performance of CNN-based methods. 
         Precision-Recall curves showing the performance obtained  
         a) after SVM classification, and b) after bounding box regression.}
\vspace{-2mm}
\label{fig:detectionExperiment}
\end{figure}

\begin{figure*}
\centering
\includegraphics[width=1\linewidth]{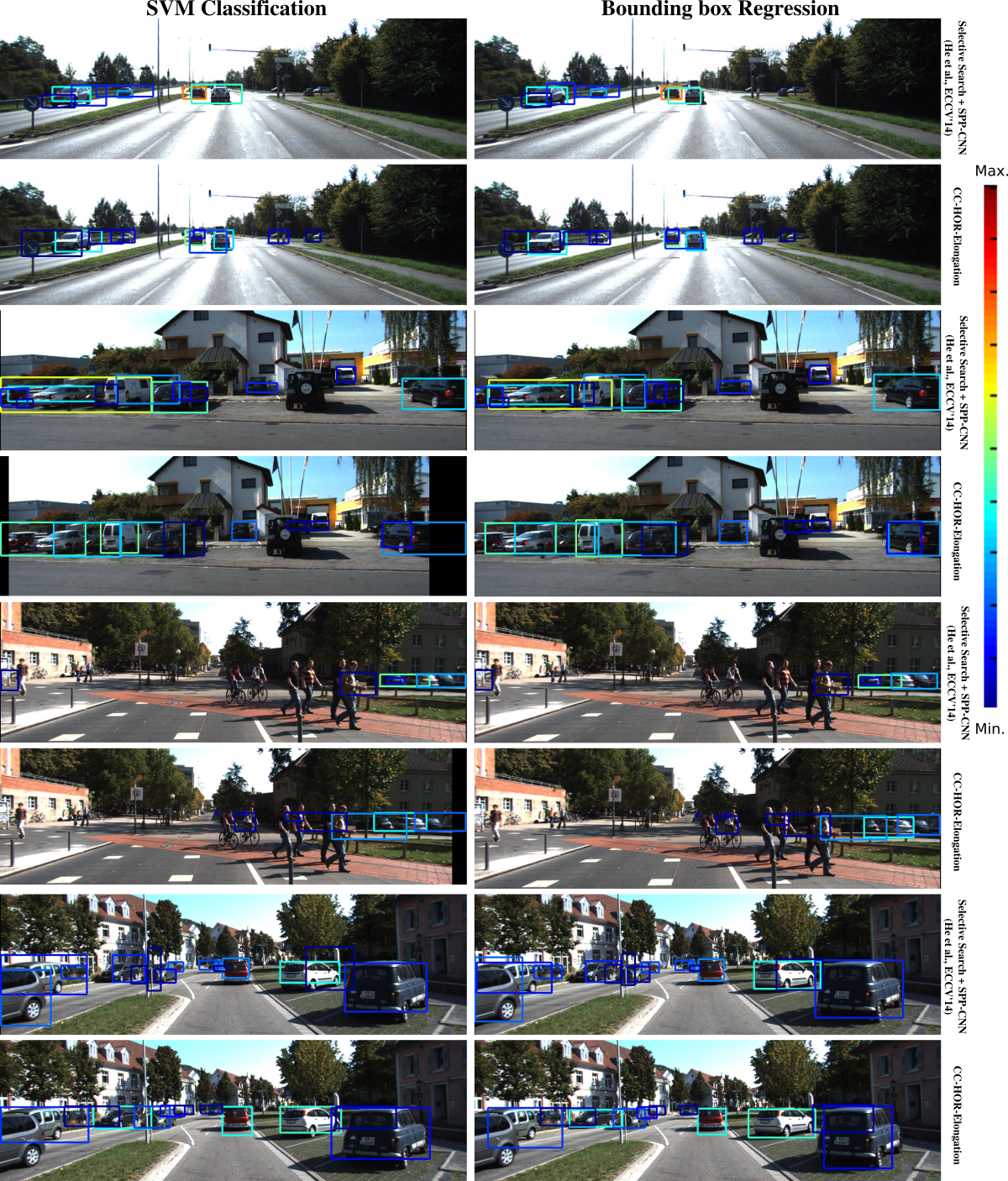}
\caption{Qualitative examples from the CNN-based methods for object detection.
         For each image the detection score of each hypothesis is color-coded 
         in jet scale.
         We show the examples for the methods from \cite{HeSPPCNN} 
        (SPP-CNN + Selective Search \cite{UijlingsIJCV2013}) and the proposed
         camera-centered higher-order relations method (CC-HOR-Elongation).
         We show the hypotheses predicted 
         a) after SVM classification, and b) after bounding box regression 
         (Best viewed in color).}
\label{fig:detectionExperimentQualitative}
\end{figure*}

While the results obtained in the previous experiments show a significant
improvement in recall, it is arguable whether the cost of decreased precision 
is acceptable.
We argue that, in systems with various sources of information (multimodal 
sensors, multi-cameras or image sequences) it is desirable to detect 
the majority of the objects since the pool of detections can be further 
reduced by imposing consistency along the different sources.  
In order to get a notion of the potential of the proposed 
methods for the object detection task, we will now follow the 
traditional object detection evaluation protocol. 
We report Mean Average Precision (mAP) as performance metric and
use the standard matching criterion from Pascal {\footnotesize VOC (IoU$>$0.5)}.
First, we report the performance for the \textit{raw} set of object 
proposals ({\footnotesize 1000} objects/image) from the previous experiments. 
Second, aiming at reducing the number of false positives per image 
and at having a comparison w.r.t. state-of-the-art detection methods,
we re-score the \textit{raw} set of objects using appearance 
features.
For this purpose, we follow the {\footnotesize R-CNN} strategy \cite{girshick14CVPR}. 
Given a set of object proposals, we compute 
{\footnotesize CNN} features for each proposal and then classify each 
region using a linear {\footnotesize SVM}. As an additional step, 
linear regression is performed in order to fix bounding box 
localization errors.
In this experiment we consider the set of objects from the 
previous experiments as the proposals to be classified. 
Finally, we follow the {\footnotesize SPP-CNN} alternative 
from \cite{HeSPPCNN} which has comparable detection performance 
to {\footnotesize R-CNN} at a fraction of processing speed. 
In addition, we split the performance of our 
{\footnotesize CNN}-based baselines showing the performance 
after {\footnotesize SVM} classification and after 
performing bounding box regression, respectively.
See Table \ref{table:detectionPerformance} for 
some quantitative results. In Figure \ref{fig:detectionExperiment} 
we present the precision-recall curves. Following the experimental 
protocol presented in this section, we present performance curves 
for SVM classification (Figure \ref{fig:detectionExperiment}.a) and 
for bounding box regression (Figure \ref{fig:detectionExperiment}.b).
See Figure \ref{fig:detectionExperimentQualitative} for some qualitative examples 
from this experiment.

\textit{Discussion:}
We notice that when the set of raw objects is considered, 
our relations-based methods lead the performance table. 
This group of methods is followed by the \textit{Relaxed NMS Detector}, 
the class-independent methods and the random methods, respectively.
\newtext{This shows that at this ``raw'' level, the proposed relations-based 
methods are better suited to cope with the variations in object 
appearance caused by occlusions and changes on scale and viewpoint.}
For the case when appearance-based re-scoring is performed, 
it is important to notice that the combination of {\footnotesize SPP-CNN}
with \textit{Selective Search} proposals corresponds to the 
state-of-the-art method proposed in \cite{HeSPPCNN}. 
Furthermore, as mentioned earlier, this is a speeded-up 
version of {\footnotesize R-CNN} \cite{girshick14CVPR}.
We can notice that the proposed relations-based methods 
produce an improvement of {\footnotesize $\sim$5} percentage points (pp) 
over {\footnotesize R-CNN} (Selective Search + {\footnotesize SPP-CNN}). 
Furthermore, the comparable good results based on the 
\textit{Relaxed NMS Detector} proposals suggests that using a 
weaker detector as proposal generator can boost the results 
obtained with {\footnotesize SPP-CNN} features.
While not at the core of our paper, this seems an interesting 
observation.
In addition, when comparing the difference in performance 
within the {\footnotesize SPP-CNN} setup, it is clear that our 
relations-based methods benefit more from bounding box 
refinement ({\footnotesize $\sim$13 pp} improvement) than the appearance-based 
\textit{Relaxed NMS Detector} ({\footnotesize $\sim$7 pp}). This further confirms 
our observation made in the previous experiments that 
context-based proposals are better at spotting regions 
of the image likely to contain the objects while 
appearance-based approaches are better suited for finer localization.
\newtext{
An additional difference between the performance of the 
\textit{Relaxed NMS Detector} and the relations-based methods lies  
in their processing times. In their current state, the evaluated
baselines, e.g. \textit{Relaxed NMS Detector} and the relations-based 
methods, have a processing bottleneck in the way in which the seed 
hypotheses are obtained. Since for the \textit{Relaxed NMS Detector} the 
set of hypotheses is high ($\sim$1000 hypotheses), its processing 
time is much higher than the relations-based methods which usually 
start from a set of $\sim$7 seed object hypotheses. This difference 
in processing times will be further discussed in Section \ref{sec:processingTimes}.
}

\begin{table}[h!]
\small
\centering

\begin{tabular}{|l|c||c|c|}

\cline{3-4}
   \multicolumn{2}{c}{}   & \multicolumn{2}{|c|}{\bf SPP-CNN}   \\ \hline 
\textbf{Baseline} & \bf{Raw} & {SVM class.} & {bbox regress.}  \\ \hline 
3D Sliding Window       & 31.30 	& 21.11		& 33.53  \\	
Random 2D Windows  	& 30.97		& 18.00 	& 24.63  \\
\hline
Selective Search	& 31.40 	& 33.50 	& 43.50  \\
Edge Boxes		& 31.29 	& 35.12 	& 39.63  \\
\hline
CC-Pairwise		& 36.37 	& 21.44 	& 31.19  \\
CC-HOR			& \bf{37.68} 	& 34.20 	& \bf{48.19}  \\
CC-HOR-Elongation		& 36.25 	& 34.30 	& \bf{48.40}  \\

\hline
Relaxed NMS Detector	& 34.40 	& \bf{42.77} 	& \bf{48.27}  \\

\hline 
\end{tabular}
\caption{\small Detection performance. Mean Average Precision (mAP).} 
\vspace{-2mm}
\label{table:detectionPerformance}
\end{table}

\subsection*{Exp.6: Comparison w.r.t. the State-of-the-Art}
\begin{figure}[ht!]
\centering
\includegraphics[width=0.92\linewidth]{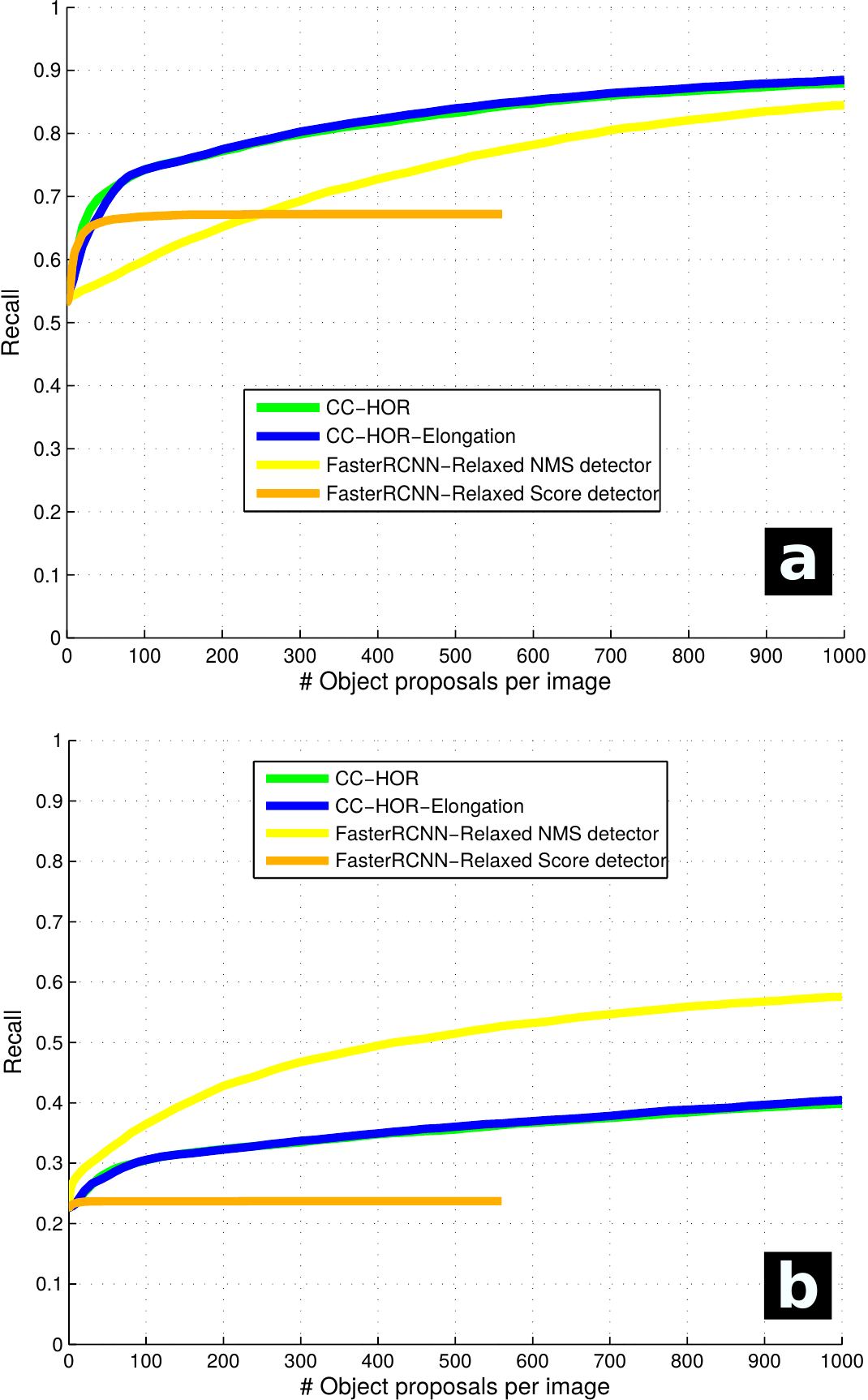}
\caption{Recall vs. number of generated proposals for our Relations-based methods 
when seed objects are collected with the Faster RCNN detector~\cite{ren15fasterrcnn} 
at default settings. For reference, we also report the performance of the 
\textit{Relaxed Score/NMS} detectors.
We report performance when using : a) a traditional matching 
criterion (IoU$>$0.5), and b) using a stricter matching criterion (IoU$>$0.75).}
\vspace{-2mm}
\label{fig:fastRCNNexp}
\end{figure}

\newtext{
In order to further evaluate the strength of our context-based 
methods at recovering missed object instances, in this experiment 
we evaluate its performance when starting from a state-of-the-art method 
for object detection, i.e. the Faster R-CNN detector \cite{ren15fasterrcnn}. 
In this experiment we define three methods, the \textit{Vanilla Faster 
R-CNN detector} which is the the original version of the detector as 
proposed in \cite{ren15fasterrcnn}, with default parameters for detection 
score threshold and non-maximum suppression. As second set of method s
we have the relaxed versions (\textit{Relaxed Score} and \textit{Relaxed NMS}) 
of the detector.
Finally, we have our context-based methods where each of the 
hypotheses produced by the \textit{Vanilla R-CNN detector} 
are enriched with viewpoint predictions using a multiclass SVM 
classifier trained from CNN features~\cite{jia2014caffe} computed 
from annotated instances in the dataset. Something important to note, 
is that this classifier is not perfect: it achieves a training cross validation accuracy of 0.4. 
However, its performance is above chance levels so it can give us an idea 
of the viewpoint (or at least the elongation angle) of an object.   
Similar to the experiments reported earlier, we used the hypotheses 
collected by the Faster RCNN detector, with default settings, as seed objects.
Based on these hypotheses we sample context-based object proposals.
Similar to the previous experiments, for the case when no seed objects 
are available, i.e. images where the object detector was unable to find 
detections above the threshold (1$\%$ of the images), we fallback to 
the \textit{3D Sliding Window} strategy and consider the proposals 
proposed by this strategy for that image.
We report performance in terms of Recall as a function of the number 
of sampled object proposals (see Figure~\ref{fig:fastRCNNexp}).
}

\newtext{
\indent\textit{Discussion:}
Clearly, the Faster RCNN detector achieves better performance 
(achieving an initial Recall $\sim$0.55) than the DPM-based detector 
used earlier in our experiments (which achieved an initial Recall 
$\sim$0.35). As a result, the proposed context-based methods are fed 
with better seed objects resulting in a boost in performance (now 
being able to reach a recall of $\sim$0.9). Note that, as was stressed 
earlier, this is achieved by using noisy object viewpoint estimates.  
We believe that a state-of-the-art method for viewpoint estimation, 
e.g. \cite{Choy_2015_CVPR}, can help to boost the performance of the 
proposed context-based methods further. 
In this experiment we also note the same trend on the performance of 
the different methods when using a stricter intersection over union 
(IoU) matching criterion (Figure~\ref{fig:fastRCNNexp}.b). 
}

\section{Processing times}
\label{sec:processingTimes}
\newtext{
Regarding processing times, each of the methods in Table~\ref{table:detectionPerformance} 
considers the same number of proposals/image. This leads to similar
processing times during the appearance-based re-scoring (classification/regression).
Hence, the difference in the processing time between the evaluated 
methods is determined by their respective methods to generate object proposals.
As mentioned earlier, the proposal generation process of the proposed 
method consists of two steps: a) class-specific object seed detection, and 
b) context-based proposal generation.
} 

\newtext{
In its current state, the bottleneck of the proposed method lies in the 
seed detection step which is handled by an off-the-shelf detector \cite{GeigerNIPS11}. 
Class-specific object seed detection takes on average 20 seconds/image 
when using the detector at default settings.
}
\newtext{
For the second step, given that the context models have been 
computed offline (Section~\ref{sec:objectProposalGeneration}), 
the execution of the proposed method can be summarized into three main processes, 
i.e. 2D-3D projection, topic assignment and object proposal sampling; which all  
scale linearly w.r.t. the number of desired object proposals. 
These processes take approximately 0.5, 0.1 and 1 seconds/image, respectively, 
giving a total of 1.6 seconds/image for the sampling of context-based proposals.
}

\newtext{
For the case of the \textit{Relaxed NMS Detector}, the object detector needs
to be evaluated for a large set of windows. 
Note that compared to the proposed method, the number of hypotheses collected 
by the off-the-shelf detector is very high in the \textit{Relaxed NMS Detector}. 
This increases the processing within the off-the-shelf detector \cite{GeigerNIPS11} 
to 30 seconds/image and further increases the bottleneck mentioned earlier. 
}

\newtext{
Note that the computation times presented above are obtained by performing 
only CPU-based computations. Moreover, for both the proposed method and 
the \textit{Relaxed NMS Detector}, this problem can be alleviated by using 
faster detectors, e.g. \cite{Benenson2012Cvpr,HeSPPCNN,ren15fasterrcnn}. 
This is supported by our experiments based on the Faster RCNN detector~\cite{ren15fasterrcnn}
( Section~\ref{sec:evaluation}, Exp.6) where on the one hand the detection 
of seed objects takes  $\sim$0.17 seconds/image while on the other hand 
the \textit{relaxed NMS} detector takes $\sim$1.12 seconds/image. 
}

\section{Limitations and Future Work}
\label{sec:limitations}
\newtext{
Even though we have shown that the proposed method is effective 
at recovering object instances missed after an initial detection 
step, there are several aspects in which the proposed method 
can be improved. In this section we look at these weak 
points and suggest directions for addressing them in future work.
%
%
\newtextrev{
Currently, our evaluation is focused on grounded objects of a single-class, 
i.e. the car class on a groundplane. 
Even though, our relations-based models can be extended to cover other 
classes not necessarily on the groundplane, e.g. by 
adding the relative {\small $Y$} location {\small $r_Y$} and related object 
class {\small $r_C$} as relation attributes, making this extension comes with 
the cost of requiring additional training data. As was presented in Section~\ref{sec:contextBasedProposals}, 
the proposed methods to generate context-based proposals learn relations between objects 
from training data. Thus, as the definition of pairwise relations gets more complex, 
more representative training data would be required in order to cover all the new 
scenarios that might be possible with the new extended pairwise relations model.
In this regard, further experiments should be performed to verify the performance 
of these extended models.
}
In addition, by being class-specific, our method may not scale 
properly if a large number of object classes need to be detected. 
In this regard, we suggest the usage of our method for structured 
scenarios with a reduced number of classes, e.g. autonomous driving 
and indoor object detection, or as a detector for specific scene-types.
Considering a specific type of setting or scene, will reduce the number 
of object classes that need to be analyzed during test time making the 
scalability aspect less critical.}

\newtext{
Regarding the object seed detection step, in its current state, our 
method requires a detector that provides a viewpoint as part of its output.
This requirement can be alleviated by modifying the relations-based models 
to focus on spatial relations (ignoring the relative pose information). 
This change comes at the cost of less interpretable context models.
A more promissing solution follows the recent line of work from \cite{Choy_2015_CVPR, RematasArxiv2016} 
which focuses on registering 3D (CAD) models to objects depicted in 2D images. 
As is presented in \cite{Choy_2015_CVPR}, this registration can be successfully exploited 
to enrich detected object hypotheses (bounding boxes) with information related 
to viewpoint. Given the increasing amount of 3D models 
appearing everyday, methods like \cite{Choy_2015_CVPR} clearly address 
the requirement of having object detections with predicted viewpoint. 
Moreover, as was presented on Exp.6 (Section~\ref{sec:evaluation}), 
even when using a relatively simple, and noisy, viewpoint estimator~(CNN featuress+SVM) 
decent performance can be achieved by the proposed method.
}

\newtext{
As presented in Figure~\ref{fig:resultsPlots}.d, when focusing 
on fine object localization, using an exhaustive dense \textit{Relaxed NMS Detector} 
outperforms the proposed method. In order to improve the performance 
of the proposed method on the fine localization task, inspired by \cite{long2013active}, 
we propose to follow a top-down approach in which given a set of object seeds 
we generate a set of initial relations-based proposals from which proposals 
with controlled variations are sampled. Size and location of these additional 
proposals are ruled by statistical data related to the class of interest and 
the spatial location of other objects in the scene.
}

\newtext{
Regarding the assumptions made on the proposed method, having a calibrated camera 
might sound as a strong assumption. However, note that existing works 
\cite{Bao2011,HoiemPOPCVPR06,Wang2005311,Marta2001} have proposed several methods 
to perform this calibration.
}

\newtext{ 
In this work we have focused our evaluation on the KITTI 
dataset~\cite{Geiger12KITTI}. As stated in Section~\ref{sec:evaluation}, 
this dataset constitutes a perfect testbed for our analysis since it covers a wide
variety of difficult scenarios, e.g. object instances with high occlusions, object 
instances with very small size, etc. Furthermore, it provides precise annotations
from objects in the 2D image and in the 3D space, including their respective viewpoints
and poses. Finally, most of the images of the KITTI dataset contain more than one 
instance of the class of interest, i.e. car, which is necessary for learning the 
relations between objects. As future work, further evaluation of the proposed method 
should be performed as new datasets showing similar properties to the KITTI dataset appear.
}

\newtext{
Finally, comparing the performance of both the \textit{Relaxed Score} 
and \textit{Relaxed NMS} detectors, suggests that a proper balance between 
their thresholds, i.e. NMS and detection score threshold, can be obtained 
in order to improve detection performance. This somehow goes against the 
common practice of focusing on the detection score threshold alone and 
leaving NMS as a fixed post-processing step. Moreover, a good balance between 
the two relaxed methods may produce a better set of seed objects for our 
context-based method for generating proposals.
}

\section{Conclusions}
\label{sec:conclusions}
In this paper we have shown that sampling class-specific 
context-based object proposals is an effective way to recover 
missed detections.
The potential of our method to improve detection 
is shown by our straightforward {\small CNN} extension 
which achieves improved performance over state-of-the-art 
{\small CNN}-based methods.
Our experimental results suggest that object proposal 
generation should not be employed solely as a pre-detection step 
as it is commonly found in the literature.
Furthermore, we show relations-based strategies 
are better suited for spotting regions that contain objects 
of interest rather than achieving fine localization.
In addition, our novel method to discover higher-order 
relations is able to recover semantic 
patterns such as traffic patterns found in urban scenes.
Future work will focus on investigating the 
complementarity of the proposed strategies as well as proper 
ways to integrate them.

\section*{Acknowledgments}
This work is supported by the FWO project 
``Representations and algorithms for the captation, 
visualization and manipulation of moving 3D objects, 
subjects and scenes'', and a NVIDIA Academic 
Hardware Grant.





\newpage
\bibliographystyle{elsart-num-sort}
\bibliography{egbib}






\end{document}